# Differentiable, learnable, regionalized process-based models with multiphysical outputs can approach state-of-the-art hydrologic prediction accuracy

Dapeng Feng, Jiangtao Liu, Kathryn Lawson, and Chaopeng Shen*
Civil and Environmental Engineering, The Pennsylvania State University
* Corresponding author: cshen@engr.psu.edu

**Abstract:**
Predictions of hydrologic variables across the entire water cycle have significant value for water resource management as well as downstream applications such as ecosystem and water quality modeling. Recently, purely data-driven deep learning models like long short-term memory (LSTM) showed seemingly-insurmountable performance in modeling rainfall-runoff and other geoscientific variables, yet they cannot predict untrained physical variables and remain challenging to interpret. Here we show that differentiable, learnable, process-based models (called δ models here) can approach the performance level of LSTM for the intensively-observed variable (streamflow) with regionalized parameterization. We use a simple hydrologic model HBV as the backbone and use embedded neural networks, which can only be trained in a differentiable programming framework, to parameterize, enhance, or replace the process-based model modules. Without using an ensemble or post-processor, δ models can obtain a median Nash Sutcliffe efficiency of 0.732 for 671 basins across the USA for the Daymet forcing dataset, compared to 0.748 from a state-of-the-art LSTM model with the same setup. For another forcing dataset, the difference is even smaller: 0.715 vs. 0.722. Meanwhile, the resulting learnable process-based models can output a full set of untrained variables, e.g., soil and groundwater storage, snowpack, evapotranspiration, and baseflow, and later be constrained by their observations. Both simulated evapotranspiration and fraction of discharge from baseflow agreed decently with alternative estimates. The general framework can work with models with various process complexity and opens up the path for learning physics from big data.

**Plain language summary:**
Recently, deep neural networks like long short-term memory (LSTM) have received a lot of attention for producing high-accuracy simulations in hydrology, but they do not respect physical laws and remain difficult to understand. However, what if you can have a model with similar accuracy, but with clarity about physical processes? What if at the same time the model respects physical laws like mass conservation and produces interpretable outputs (like soil moisture, groundwater storage, and baseflow), with which you can tell a whole story to stakeholders? What if the same framework allows you to ask precise questions about different parts of hydrology and re-examine your understanding of some parts of the physical system or check if your past equations are correct? This paper delivers a system that achieves these grand goals and opens many avenues for further exploration.

## 1. Introduction:

Regional hydrologic models have been widely deployed for operational flood forecasting (Johnson et al., 2019; Maidment, 2017), future change projection (Hagemann et al., 2013), water resources management (Beck et al., 2020; Guo et al., 2021; Mizukami et al., 2017), and supporting downstream applications such as crop (Ines et al., 2013) and water quality modeling (Dick et al., 2016; Strauch et al., 2017). The accuracy of these models has important implications for relevant government agencies and public stakeholders that place trust in them. The demand for accurate modeling capabilities will likely be on the rise due to increased risks of floods and droughts because of climate change (IPCC, 2021).

Traditionally, regional hydrologic models describe not only streamflow but also other water stores in the hydrologic cycle (snow, surface ponding, soil moisture, groundwater), as well as fluxes (evapotranspiration, surface runoff, subsurface runoff, baseflow). The physical states (stores) and fluxes help to provide a full narrative of the event, e.g., high antecedent soil moisture or thawing snow primed the watershed for floods, which are important for communication with stakeholders. It allows us to ask specific scientific questions like identifying and characterizing types of floods (Berghuijs et al., 2016). Some process granularity (meaning the ability to describe the level of details that discern different processes) is also critically important for downstream applications. For example, agricultural models require knowledge of soil moisture and evapotranspiration demand, while water temperature models require a separation between surface runoff and baseflow. Through calibrating the models to streamflow, the hope was that the other fluxes and states would also be constrained via their physical linkages, but this is complicated due to parameter non-uniqueness.

Recently, purely data-driven deep learning (DL) models (LeCun et al., 2015; Shen, 2018a, 2018b) showed surprisingly strong performance in hydrologic modeling, but they do not resolve internal hydrologic dynamics. These models have very generic internal structures that allow them to learn directly from big data without invoking many problem-specific theories and assumptions. A particularly popular architecture in hydrology is long short-term memory (LSTM) (Hochreiter & Schmidhuber, 1997). LSTM's accuracy has been demonstrated for many hydrologic variables on large datasets including soil moisture (Fang et al., 2017, 2019; Fang & Shen, 2020; Liu et al., 2022), streamflow (Konapala et al., 2020; Kratzert, Klotz, Shalev, et al., 2019; Sun et al., 2021; Xiang et al., 2020; Xiang & Demir, 2020), dissolved oxygen (Zhi et al., 2021), evapotranspiration (Zhao et al., 2019), groundwater (Sahu et al., 2020; Solgi et al., 2021; Wunsch et al., 2021), and water temperature (Rahmani, Lawson, et al., 2021; Rahmani, Shen, et al., 2021), covering every part of the hydrologic cycle (Shen et al., 2021). It is widely publicized that LSTM represented a "step-change" in performance which also suggests our traditional models were far from optimality (Nearing et al., 2021). The growth from 2 LSTM papers in hydrology in 2017 to 300+ papers in 2021 (Shen & Lawson, 2021) demonstrated its appeal and popularity.

Nevertheless, many variables of interest are not adequately observed, so pure DL models do not apply to them. For hydrologic modeling, it remains challenging to interpret what is being learned by LSTM, in part because it does not output physical internal states and fluxes. While it is possible to correlate LSTM's cell states to some physical states, there is no guarantee of the meaning of these states, and the physical connections between the states are unclear. It also does not allow us to freely ask questions about how the system functions. The hydrologic community appears to be at a crossroad: seemingly, they cannot have both predictive power and scientific understanding at the same time, yet both are crucial for projecting the impacts of our future climate.

For streamflow prediction, the current best performance with LSTM without using an ensemble and with forcing data from the North American Land Data Assimilation System (NLDAS; different forcing datasets have a moderate influence on results) shows a median Nash-Sutcliffe model efficiency coefficient (NSE) of 0.72, reported for the Catchment Attributes and Meteorology for Large-sample Studies (CAMELS) dataset (Addor et al., 2017; Newman et al., 2014) with 671 basins across the USA (Feng et al., 2020; Kratzert et al., 2020). In a more interpretable modeling framework, Jiang et al. (2020) showed the possibility to connect a convolutional neural network (CNN) as a post-processor (or correction) layer to a conceptual process-based model encoded as a recurrent neural network. Their model achieved median NSE values of 0.48 and 0.71 without and with the CNN post-processor, respectively, for 591 basins (a subset of the 671 CAMELS basins). While this work is encouraging and is an important step forward, a workflow without the use of a post-processing layer could potentially provide better physical significance and interpretability.

In hydrologic modeling, there is a large distinction between locally-calibrated models and regionalized models (meaning parameters are related to autocorrelated, widely applicable features, not just calibrated on a gauge) (Hrachowitz et al., 2013). Only regionalized models are applicable to ungauged basins, though they will always have poorer performance than in gauged basins with site-by-site calibration, sometimes substantially (Beck et al., 2020; Hogue et al., 2005; Kumar et al., 2013; Rosero et al., 2010). LSTM can be regionalized when given basin-averaged attributes like soil composition or slope as inputs to distinguish between basins (in some studies, when the LSTM network is trained on a basin-by-basin basis, it essentially is a locally-calibrated model). LSTM is also not immune to performance degradation when applied to ungauged basins, but it has been shown to score higher than locally-calibrated hydrologic models even when tested out of sample (Feng et al., 2021; Kratzert, Klotz, Herrnegger, et al., 2019). In fact, regionalized LSTM remained competitive even in the scenario of being applied in large, contiguous regions with no data if some ensemble methods were used (Feng et al., 2021). While locally-calibrated models are certainly useful, in this work we focus on regionalized frameworks for wider applicability in the future.

Our earlier work showed that a framework called differentiable parameter learning (dPL) could employ big data and machine learning approaches to find parameter sets for process-based geoscientific models (Tsai et al., 2021). dPL provided similar performance compared to evolutionary algorithms for the main calibration variable, and presented better results for spatial generalization and uncalibrated variables, while also using orders-of-magnitude lower computational effort. dPL superseded earlier regionalization schemes in PUB tests. Such strengths are partly because dPL leverages a beneficial data scaling curve where the inclusion of many sites allows them to benefit each other. dPL is named as such because it applies differentiable programming (Baydin et al., 2018), which enables us to track the gradients of the outputs of process-based models with respect to input parameters or neural network weights, for the models' parameterization. Differentiability critically enables the training of large-scale neural networks to work with process-based models. While dPL gives an initial glimpse at the power of applying differentiable programming to hydrology, it alone does not penetrate model process descriptions, and thus the performance is still limited by the structure of the existing models. In terms of accuracy, dPL can find close-to-optimal parameters for process-based models, but the end outcomes still lag far behind pure DL models.

Expanding from the advances of dPL, here we demonstrate a new hydrologic modeling paradigm where learnable, differentiable process-based models with embedded neural networks (NN) can achieve similar predictive performance as LSTM models. We call them δ models because they fully leverage differentiable programming so the models are "learnable" and "evolvable". Further, such a learnable model respects mass balances, can output important internal physical fluxes, and allows us to ask new scientific questions. We imposed several additional requirements on our framework: (i) each step of the main model calculation either has physical logic associated with physical terms, or uses a neural network but with its effects quantified, (ii) mass balances are observed, and (iii) multiple internal fluxes and states are described (e.g., groundwater and surface water contributions should be distinguished in our test case). This paper is intended to be a concise report on the potential performance, methodology, and implications, with myriad research questions set in motion for the future.

## 2. Methods and datasets

As an overview, we use an existing process-based model as a backbone, which can be based on either discrete (modifying from a backbone with discrete formulation) or continuous (written in a differential equation format) formulations. Only the discrete version is demonstrated in this paper. In this example, we selected the Hydrologiska Byråns Vattenbalansavdelning (HBV) model (Aghakouchak & Habib, 2010; Beck et al., 2020; Bergström, 1976, 1992; Seibert & Vis, 2012) as the backbone. HBV is a simple bucket-type hydrologic model that simulates hydrologic variables including snow water equivalent, soil water, groundwater storage, evapotranspiration, quick flow, baseflow, and total streamflow. We then altered parts of the backbone model structure, coupled it with differentiable parameter learning (dPL) as a regionalized parameterization scheme, and replaced parts of the model with neural networks. We then trained and tested the model on the CAMELS dataset.

### 2.1. dPL and dPL+HBV for comparison

Differentiable parameter learning (dPL) only concerns the parameter space and is used to support parameterization of the evolved HBV model, as well as to serve as a comparison case when coupled with the unmodified HBV model. The dPL framework can be described concisely as θ = $g_A(A,x)$ where θ represents some parameters, *A* contains some static attributes such as topography, soil texture, land cover, and geology, *x* is the meteorological forcings (Section 2.4.1), and $g_A$ is the parameter estimation neural network. We tried treating parts of θ as either static or dynamic. If a parameter is treated as static, the same value is used throughout the HBV simulation. If it is treated as dynamic, the model gets a new value for this parameter every day, which we call dynamical parameterization (DP), denoted by superscript $^t$. The original HBV model only has time-constant parameters so we made a change to reflect the impacts of vegetation, seasonal water storage, and to correct errors with the ET formula (Section 2.2). To support dynamic parameterization, the network $g_A$ is in fact an LSTM model. A main feature of the dPL framework is that *θ* is subsequently provided to the HBV model and $g_A$ is trained together with the rest of the model by the global loss function involving all sites. In machine learning language, this framework is called "end-to-end", in that there are no intermediate ground-truth data as supervising data for *θ*. This avoids two issues: first is that we do not normally have ground-truth data for *θ* (while some soil parameters may be occasionally available, most other parameters like groundwater recession parameters or runoff factors are not); second is that we minimize the chance for error with the intermediate steps.

## 2.2. The differentiable, learnable process-based model

The time-discrete HBV model is described succinctly in Table 1 and illustrated in Figure 1. This model has five state variables. We implemented the time-discrete HBV model on PyTorch, a platform supporting automatic differentiation (Paszke et al., 2017). Other platforms could work similarly.

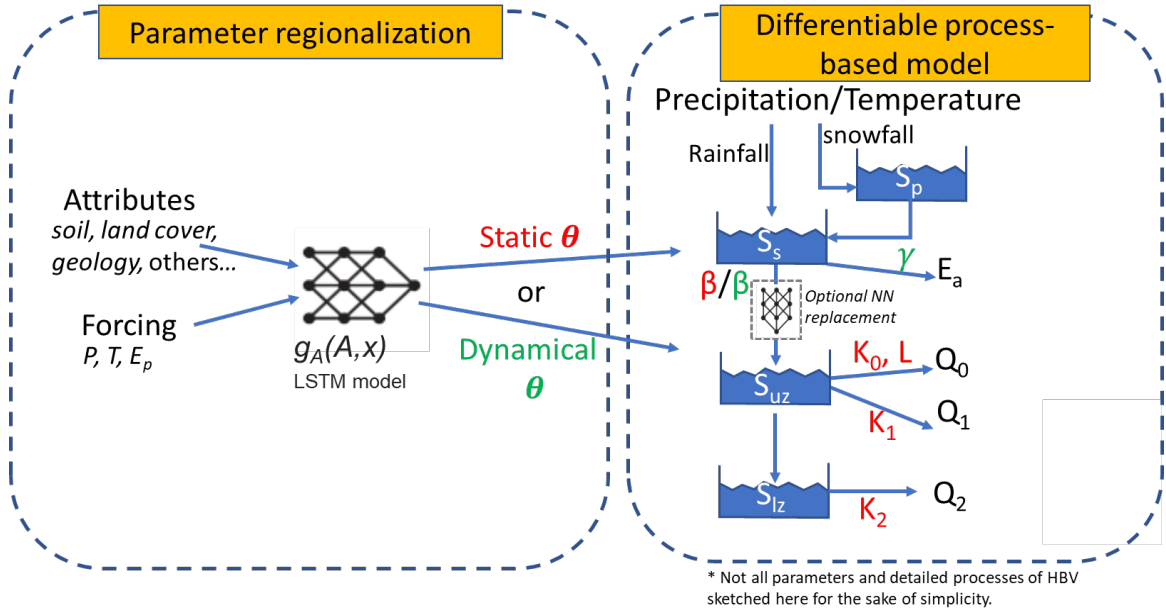

*Figure 1. The sketch of the evolved differentiable parameter learning (dPL) to the HBV model. Red symbols are static parameters and green symbols are time-dependent parameters.*

*Table 1. Major equations for the original and modified time-discrete HBV models. For simplicity, some minor equations and thresholding functions (for preserving mass positivity) are omitted here. Δ indicates changes in a daily time step; S terms with subscripts indicate state variables; θ terms are parameters (except for β and γ which are written directly). Text explanations in square brackets.*

| Module | Default HBV equations | Equations modified |
|---|---|---|
| Parameters | [The parameter set $\{\theta,\gamma,\beta\}$ is to be calibrated individually for each basin or regionalized.] | [Regionalized calibration using dPL]<br>Static parameters:<br>$\{\theta,\gamma,\beta\} = g_A(A,x)$<br><br>Dynamic parameters:<br>$\{\theta,\gamma^t,\beta^t\} = g_A(A,x)$<br>[or only one of γ and β is dynamical] |
| Snow ($S_p$ & $S_{liq}$) | $\Delta S_p = P_s + R_{fz} - s_{melt}$ [solid snowpack]<br>--- $P_s$ [precipitation as snow]<br>--- $R_{fz} = (T_t-T)\theta_{DD}\theta_{rfz}$ [refreeze of ponding water]<br>--- $s_{melt} = (T-T_t)\theta_{DD}$ [snowmelt]<br>$\Delta S_{liq} = s_{melt} - R_{fz} - I_{snow}$ [liquid in snow]<br>--- $I_{snow} = S_{liq} - \theta_{CWH} * S_p$ [snowmelt infiltration] | unchanged |
| Surface soil water ($S_s$) | $\Delta S_s = I_{snow} + P_r - P_{eff} - E_x - E_T$ [soil water]<br>--- $P_r$ [precipitation as rain]<br>--- $P_{eff} = W(P_r + I_{snow})$ [effective rainfall to | Static parameters:<br>$\eta = \min((S_s/(\theta_{FC}\theta_{LP}))^\gamma, 1)$ |
| | produce runoff]<br>--- W= min($(S_s/\theta_{FC})^\beta$, 1) [soil wetness factor, $\theta_{FC}$ is field capacity]<br>--- $E_x$= ($S_s$-$\theta_{FC}$)   [excess]<br>--- $E_T$= $\eta * E_p$   [actual ET, $E_p$ is potential ET]<br>--- $\eta$= min($S_s/(\theta_{FC}\theta_{LP})$,1) [ET efficiency] | Dynamic parameters:<br>W= min($(S_s/\theta_{FC})^{\beta^t}$, 1)<br>$\eta$= min($(S_s/(\theta_{FC}\theta_{LP}))^{\gamma^t}$,1)<br><br>NN replacement option:<br>$P_{eff}$ = $NN_r(\theta_{FC}, \beta, S_s, S_s/\theta_{FC}, P_r+I_{snow})$ |
| Upper subsurface zone ($S_{uz}$) | $\Delta S_{uz}=P_{eff}+E_x-P_{erc}-Q_0-Q_1$ [upper subsurface zone]<br>--- $P_{erc}$ = min($\theta_{perc}, S_{uz}$) [Percolation]<br><br>--- $Q_0$= $\theta_{K0}(S_{uz}-\theta_{uzl})$   [fast flow]<br>--- $Q_1$= $\theta_{K1}S_{uz}$   [subsurface stormflow] | unchanged |
| Lower subsurface zone ($S_{lz}$) | $\Delta S_{lz}=P_{erc}-Q_2$   [lower subsurface zone]<br>--- $Q_2$= $\theta_{K2}S_{lz}$   [baseflow] | unchanged |
| Discharge | Q = $Q_0$ + $Q_1$ + $Q_1$ | |
| Routing | $Q^*(t) = \int_o^{tmax} \xi(s:\theta_a, \theta_\tau) * R(t-s)ds$<br>--- $\xi(t:\theta_a, \theta_\tau) = \frac{1}{\Gamma(\theta_a)\theta_\tau^{\theta_a}} t^{\theta_a-1} e^{-\frac{t}{\theta_\tau}}$ | |

Here are some of the explanations and rationales for the modifications:

A. We added a parameter to the ET equation (γ in Table 1's "Equations modified" column for surface soil).

B. We tested allowing γ to be dynamic ($\gamma^t$) to mitigate the errors from the potential ET equation and to mimic the impacts of vegetation. For the latter, the impacts would be season-dependent due to the phenological cycle of vegetation, e.g., in the summer, the vegetation roots will be more active; after drought, vegetation may need time to recover its water use efficiency. Because a model like LSTM can automatically capture such effects while the original HBV cannot, HBV may never reach the performance of LSTM without considering seasonally-varying parameters. We also tested setting the runoff curve parameter (β) to be dynamical ($\beta^t$) as a comparison scenario to understand the influence of these choices. This parameter characterizes the relationship between surface soil moisture and effective rainfall (the amount of water available for runoff). The curve roughly characterizes how soil wetness translates into effective rainfall, and β is inversely related to runoff. Having a dynamical $β^t$ can allow the forcing history to influence runoff production.

C. We tested replacing the effective rainfall parts of the model with a neural network to examine if there are more suitable relationships to describe the moisture-runoff relationship than the original one. The NN replacement method can be similarly applied to other modules like groundwater and ET, but here we only applied it to the effective rainfall component to constrain the model's flexibility. Other replacements can be investigated in future work.

D. As in Tsai et al. (2021) and Mizukami et al. (2017), we added a routing module to the hydrologic model with two parameters (Table 1). This module assumes a gamma function for the unit hydrograph and convolves the unit hydrograph with the runoff to produce the final streamflow output, which is compared to the observed streamflow. This routing module is employed in all of our HBV-based simulations in this paper.

We use the symbology $δ$ to represent different dPL+ evolved HBV models for easy reference. $δ(β^t)$ represents using both the evolved HBV and dynamic parameter $β$. We could use $δ^{HBV}$ to indicate that HBV is the backbone, and many other backbones can be used, but since all of the models in this paper used HBV as the backbone, we omitted the superscript and simply used $δ$ instead.

2.3.3 Model training, hyperparameters, and other details

We trained the models on 15 years' worth of data from 1 October 1980 to 30 September 1995 and evaluated the performance on another 15 years' worth of data from 1 October 1995 to 30 September 2010. The 5 years' worth of data from 1 October 1980 to 30 September 1985 was used to select the hyperparameters. For the hyperparameters of the LSTM model used as $g_A$ in dPL+HBV (Figure 1), we manually tested several combinations and chose 256 hidden states. A mini-batch size of 100 and a training length of 365 days were used to train the dPL+HBV framework. Additionally, we used one year of meteorological forcings as the warm-up period for initializing the state variables of HBV during training. Except for the dPL models, we also ran the purely data-driven LSTM streamflow model on the same time periods for performance comparison. The setup of the LSTM streamflow model was the same as in our previous study (Feng et al., 2020). Additionally, to be comparable with previous regionalized modeling studies (Rakovec et al., 2019), we also run some dPL experiments on the same training/testing periods as theirs.

2.3.4 Loss function and evaluation metrics.

Our loss function (Equation 1) was defined as a weighted combination of two parts based on root-mean-square error (RMSE) calculations on all basins (in practice, a minibatch of basins during training). The first part was the RMSE calculated on the predicted and observed streamflow, while the second part was the RMSE calculated on the transformed streamflow (Equation 2). The transformation in the second part of the loss aims at improving low flow representation. We used a parameter $\alpha$ to assign weights to different parts of the loss function and the value 0.25 was used here to train all the HBV models. For the benchmark LSTM model, we applied the transformation (Equation 2) and normalization as preprocessing steps, which were found to give slightly better NSEs in our previous study (Feng et al., 2020). The loss function of LSTM was the RMSE of the normalized data.

$$Loss = (1.0 - \alpha)\sqrt{\frac{\sum_{b=1}^{B}\sum_{t=1}^{T}(Q_S^{t,b} - Q_O^{t,b})^2}{B*T}} + \alpha\sqrt{\frac{\sum_{b=1}^{B}\sum_{t=1}^{T}(\hat{Q}_S^{t,b} - \hat{Q}_O^{t,b})^2}{B*T}} \quad (1)$$

$$\hat{Q}^{t,b} = Log_{10}(\sqrt{Q^{t,b}} + 0.1) \quad (2)$$

Here $Q^{t,b}$ represents the streamflow on day *t* and basin *b*; the subscripts *s* and *o* represent simulations and observations, respectively; $\hat{Q}^{t,b}$ is the transformed streamflow in order to improve low flow representations; *B* and *T* respectively represent the number of basins (same as the batch size) and total days (same as the length of training instance) in a training minibatch; and $\alpha$ is a weighted parameter which is 0.25. For model evaluation, we computed Nash-Sutcliffe model efficiency coefficients (NSE). NSE is one minus the ratio of the error variance of the modeled time series divided by the variance of the observed time series. It is 1 for a perfect model and 0 for the long-term mean value used as the prediction.

**2.4. Input and observation datasets**

2.4.1. Forcing, streamflow, and attribute data

We used the Catchment Attributes and Meteorology for Large-sample Studies (CAMELS) dataset (Addor et al., 2017; Newman et al., 2014) which contains meteorological forcings, streamflow observations, and attributes for 671 basins in the conterminous United States (CONUS). The North American Land Data Assimilation System (NLDAS) (Xia et al., 2012) meteorological forcings from CAMELS were used to run the experiments, with the daily minimum and maximum NLDAS temperature data obtained from Kratzert (2019). The HBV model only used three forcing variables: precipitation (P), temperature (T), and potential evapotranspiration ($E_p$). Describing the total evaporative demand, $E_p$ was estimated using the Hargreaves (1994) method, which considers mean, maximum, minimum temperatures and latitudes. For $g_A$, x={P, T, $E_p$} was used as the dynamical forcings. For the comparison LSTM model, we used {P, T, solar radiation, vapor pressure, day lengths} as forcings. The inputs to HBV is pre-determined by its structure while it can be any relevant variable for LSTM.

We used the streamflow data compiled by the CAMELS dataset, which was, in turn, obtained from the streamflow network of the United States Geological Survey (USGS). Static attribute data from CAMELS including 35 topography, climate, land cover, soil, and geology variables in total (Table A1 in the Appendix) were included as inputs to dPL to train regionalized models. We also acquired the simulations of other regionalized models from previous studies as comparisons for our dPL models (Kratzert, Klotz, Shalev, et al., 2019; Rakovec et al., 2019).

2.4.2. Baseflow index and evapotranspiration for evaluation

The baseflow index ($BFI^{L13}$) was derived from applying Lyne and Hollick filters with warmup periods to streamflow hydrographs by Ladson et al, (2013), which was compiled and included in CAMELS. Moderate Resolution Imaging Spectroradiometer (MODIS), part of the National Aeronautics and Space Administration Earth Observing System (NASA/EOS) project, uses satellite data to estimate terrestrial surface evapotranspiration. The improved ET algorithm of the MOD16-MT dataset is based on the Penman-Monteith equation (Monteith, 1965; Mu et al., 2011). The input data includes meteorological reanalysis data of daily surface downward solar radiation and air temperature, and MODIS products such as albedo, land cover, leaf area index, and fraction of photosynthetically active radiation. The data cover all basins in CAMELS. We composited the outputs from our models as 8-day data following the same composite methodology as MODIS, and compared them.

## 3. Results and Discussion

We first briefly showcase the surprising performance of the evolved differentiable HBV models and put them in the context of previous literature. Then, we compare simulated internal variables to other estimates.

### 3.1. Streamflow metrics.

On the CAMELS dataset, strikingly, some of the δ models achieved comparable Nash-Sutcliffe model efficiency coefficient values (NSE=0.715) to LSTM (NSE=0.72 with NLDAS without ensemble -- this result is the same with either our LSTM or that of Kratzert et al., (2020) (Figure 2 and Table 2). They are both clearly ahead of dPL+HBV (representing the unmodified model with optimized, regionalized parameters, median NSE=0.628), and MPR+mHm (representing another traditional model with a different regionalization scheme, median NSE=0.53).

These strengths in metrics have strong real-world implications as shown by time series comparisons for several example basins (Figure 3). The HBV model had difficulties capturing flood peaks for basin A in east Texas, while the evolved structure can mitigate this problem (Figure 3a). For basin B, close to the Great Lake, the evolved model corrected the timing error of the peak flows from the original HBV model (Figure 3b). We can clearly see an improved flow dynamics from basin C located in the Rocky mountains. The evolved model improved both the peak and baseflow estimation here (Figure 3c).

The fact that we can approach LSTM performance using a process-based, mass-conservative model is refreshing because in previous studies, the gap between LSTM and process-based models with a wide variety of structures appeared to be uncrossable. These results indicated that it is difficult for expert-derived formulas to match the ability of trained neural networks, a pattern we have seen repeatedly in recent studies across various disciplines. In this work, however, we observe that the gaps in performance between LSTM and learnable models ($\delta$ models) are not so large. This highlights the power of having learnable structures and adaptivity in the model.

Considering the nuances of the configurations and comparing it laterally with the literature, the performance of optimization in our differentiable framework (median NSE=0.715) should be close to optimum. Firstly, we have not employed a neural network as a post-processor. Previous results showed that using an LSTM as a post-processor to a hydrologic model produces the same results as LSTM itself (Frame et al., 2021). In this case, post-processors can strongly alter the simulation so that the contributions from the front-end model are no longer clear. Secondly, we have not employed an ensemble, which could elevate median NSE from 0.72 to 0.74 with NLDAS forcing (Feng et al., 2020; Kratzert, Klotz, Herrnegger, et al., 2019) and could also help traditional models (Seibert et al., 2018), and will be explored in the next stage. Thirdly, as mentioned earlier, previous results with traditional hydrologic models almost always showed regionalized parameterization to be weaker than site-by-site calibration --- previous best results with regionalized hydrologic models had a median of 0.53 (Beck et al., 2020; Rakovec et al., 2019). However, the result of our regionalized formulation is even close to the recording-holding LSTM models. Presumably, like LSTM, it also benefits from the efficiency and data synergy of learning from big data (Fang et al., 2022).

The models that do not utilize dynamical parameterization (DP) already presented substantial improvements over traditional models, but to truly approach the performance of LSTM, we needed to invoke DP. The δ model (without DP) was at median NSE values of 0.694 (Table 2), already significantly higher than regionalized HBV (NSE=0.628) or mHm (NSE=0.53). This shows that DP is not necessarily needed for the model to be operational. However, we should not let go of trying to further close the gap to LSTM (from 0.697 to 0.72) because it is symptomatic of some more fundamental issues. After various attempts, we were never able to get a median NSE above 0.705 without DP.

The estimated dynamical runoff factor $β^t$, which is negatively related to runoff, has a main seasonal cycle with spikes corresponding to storms (Figure 4, obtained from the $δ(β^t)$ model). $β^t$ starts at the highest level in October, reaches bottom in March or April, and starts to rise in May or June. It corresponds to increasing runoff (other conditions being equal) from October to March, and reducing runoff from May to September, which matches the rhythm of the seasonal water storage cycle. In the original HBV, only the surface layers ($S_s$ & $S_{uz}$) influence runoff, and the deeper groundwater reservoirs have no feedback to surface runoff, i.e., quick flow generation is not influenced by how much water there is in deeper groundwater layers in the model. In reality, this feedback can be important as large water storage in the basin can prime the basin for large runoffs. Hence, an explanation is that the dynamical parameterization captured this problem and introduced a mechanism for high water storage accumulated over months to increase runoff. This is consistent with the observations that $β^t$ in the hot Texas basins (Figures 3a & 4a) has more flashy and less seasonal response -- runoff in these basins tend to be flash infiltration excess and the effect of water storage is limited. Another concurrent possibility is $β^t$ also reflects more established canopy in warm seasons to increase abstraction and reduce runoff.

Regarding why it is useful to enable dynamical parameterization for $γ^t$, which is related to ET, we hypothesize that two factors are at play: (i) the potential ET formula (Hargreaves) could not adequately capture the ET demand but DP (with $γ^t$) could correct it; and (ii) there are some missing memory mechanisms (or states) in the design of the HBV model, in terms of representing seasonally-varying ecosystem states like vegetation density, leaf area index, rooting depth, and xylem states (Mackay et al., 2015). These states reflect the impacts of accumulated forcings at the monthly scale. Models like LSTM could implicitly capture such seasonal effects. In contrast, the original HBV model, as well as many other conceptual models, simply does not have a functional vegetation module.

For $γ^t$ and $β^t$, their states in DP reflects the impacts of accumulated forcings at the monthly scale. Models like LSTM could implicitly capture such seasonal effects. In contrast, the original HBV, as well as many conceptual models, simply does not have these processes and memory. Other parameters must be distorted for the original model to compensate for their absence. If our theory is true, we expect that no process-based models, learnable or not, can achieve optimal performance without adding these states.

Table 2. Model performances for the test period (1995-2010) with NLDAS forcing data and without the use of ensemble. δ refers to the differentiable process-based model built on HBV as the backbone; Superscript t (as in $γ^t$ or $β^t$) indicates where dynamical parameterization is applied (otherwise, parameters are treated as constants).

| Model name & explanations | Median streamflow NSE | Spatial Correlation of BFI | Median ET Correlation |
|---|---|---|---|
| $δ(γ^t, β^t)$ | 0.715 | 0.745 | 0.848 |
| $δ$ | 0.694 | 0.729 | 0.787 |
| $δ(β^t)$ | 0.711 | 0.692 | 0.790 |
| $δ(γ^t)$ | 0.712 | 0.726 | 0.833 |
| dPL+HBV | 0.628 | 0.621 | 0.788 |
| LSTM (ours) | 0.72 | - | - |
| LSTM (Table A1 in (Kratzert et al., 2020)) | 0.72 | - | - |

### 3.2. Comparisons of the baseflow index and ET

As opposed to the LSTM model which can only predict the total streamflow, the evolved HBV (δ) models can elucidate the different sources of streamflow and ET. Overall, we found that the baseflow and ET metrics tended to be consistent with NSE, that is, models with better NSE also tended to have better correlations of baseflow and ET (Table 2), although exceptions did exist. The $δ(γ^t, β^t)$ model which has the highest median NSE also has the highest baseflow correlation with $BFI^{L13}$ and median ET correlation with MODIS.

There is a decent correlation (R=0.745 for $δ(γ^t, β^t)$ and 0.729 for δ) between the simulated BFI ($Q_2/Q$) and BFI derived from Ladson et al (2013), $BFI^{L13}$ (Figure 5f), showing the subsurface module in evolved HBV has captured the rough baseflow patterns across the CONUS (Figure 5b,d). It should be noted that the results of baseflow recession analysis can vary significantly based on the procedure and assumptions employed and do not represent ground truth. Hence, we place more emphasis on the general spatial pattern rather than the absolute values in the BFI. We notice high BFI associated with thick and permeable soil in the southeast and western US, and in the Upper Colorado river basin. We also notice low BFI along the Appalachian Plateau extending into the Central Plains, in line with previous analysis of terrestrial water storage data and streamflow (Fang & Shen, 2017). Overall, the HBV-predicted BFI ($Q_2/Q$) trends towards being higher than $BFI^{L13}$. $Q_2/Q$ from basins with NSE values over 0.5 has a higher correlation with $BFI^{L13}$ than for the whole dataset. On a side note, the basins with NSE<0.5 (the basins in Figure 5a but missing in Figure 5b) concentrate on the upper Great Plains, which has been known for a long time to cause difficulties for various kinds of models (Feng et al., 2020; Newman et al., 2015). We suspect this is due to incorrect watershed boundaries, highly-concentrated rainfall, and the existence of cross-basin water transfer, i.e., large-scale groundwater flow and springs.

It is not surprising that the simulated $Q_2/Q$ agrees with the values derived from baseflow separation analysis, which is based on streamflow hydrographs. Passing water through the system and releasing it as $Q_2$ is the primary way for the model to mathematically introduce slowly-varying base flows. However, the point is that we can now diagnose different parts of streamflow contributions by learning from data, and further link them to downstream applications such as water temperature predictions.

For the original HBV, δ, and $δ(γ^t, β^t)$ models, the median temporal correlation of ET was 0.79, 0.79, and 0.85 (Table 2) and the median RMSE was 7.12, 6.77, and 6.21 mm/8days, respectively, against the 8-day integrated MOD16-MT ET product on CAMELS basins(Figure 6a). Correlation with MOD16-MT improved from the original HBV to $δ(γ^t, β^t)$, which is consistent with the improvement of streamflow NSE (Figure 6a, Table 2). These ET metrics are decent and appear better than the accuracy levels reported for other estimates evaluated against MOD16-MT. To give some context, when evaluated on a monthly scale, Velpuri et al. (2013) reported $R^2$ of ~0.56 and an RMSE between 26-32 mm/month between MOD16-MT and FLUXNET stations. Even without training on ET, the differentiable process-based models can predict reasonable daily ET, which LSTM cannot do at all.

We also note that the model with dynamic parameterization has a higher median correlation with MOD16-MT. The ET correlation of $\delta(\gamma^t)$ is better than $\delta(\beta^t)$, and $\delta(\gamma^t, \beta^t)$ is better than both of them. This suggests the system calls for memory mechanisms associated with ET. All of these observations suggest ET can be better aided by using dynamic parameterization for $\gamma$, which further suggests there are flaws with the present ET algorithm in the model, either due to imperfect potential ET estimation or lack of vegetation.

It is difficult to provide a lateral comparison based on daily ET data to the literature regarding the value of learning from streamflow. There is a lack of evaluation of ET on the CAMELS basins in the literature, while the FLUXNET data, which were often used to evaluate ET products, were not coincident with the CAMELS basins to examine the effects of learning from streamflow data. MOD16-MT, estimated using a modified Penman-Monteith equation (Mu et al., 2011), should not be interpreted as ground truth. The point here is not that this model produces more accurate ET (which will be studied in future work), but that the differentiable process-based model can now be constrained and evaluated by multi-source, multifaceted observations.

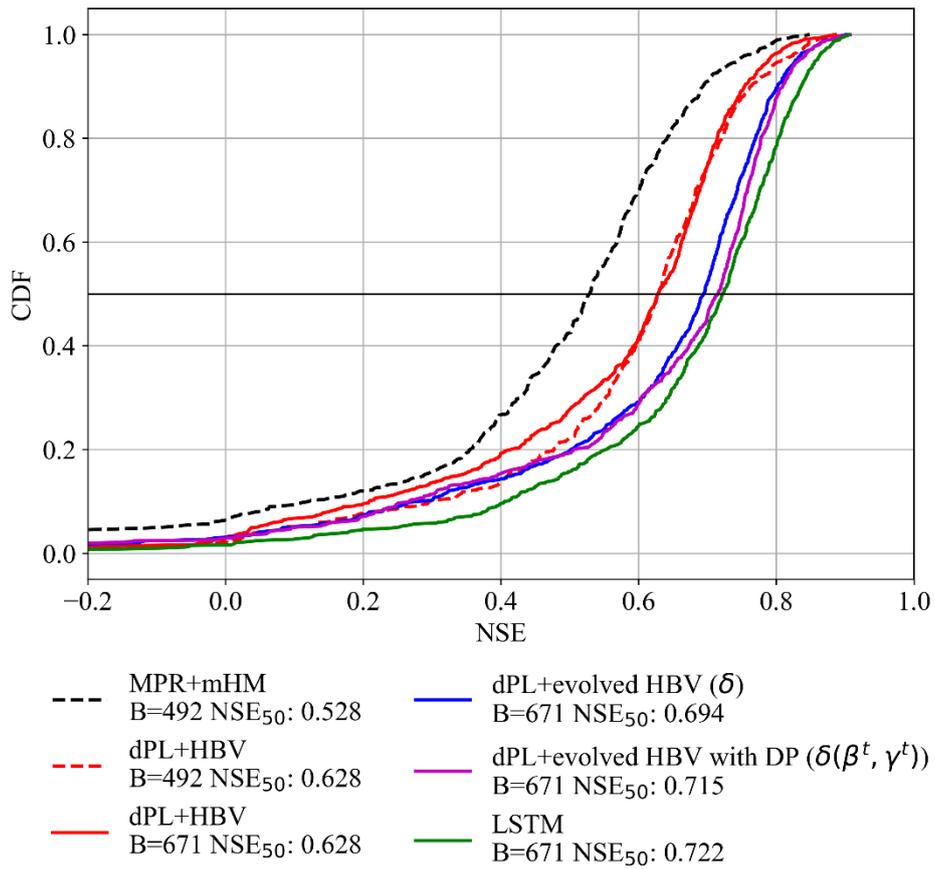

Figure 2. NSE comparison of different models on the CAMELS dataset for the testing period. The dashed lines represent models with the training/testing periods of hydrologic years 1999-2008/1989-1999, while the solid lines represent models with the training/testing periods of 1980-1995/1995-2010. MPR+mHM is originally from Rakovec et al (2019). The letter B here represents the number of CAMELS basins used for the modeling. $NSE_{50}$ is the median NSE value.

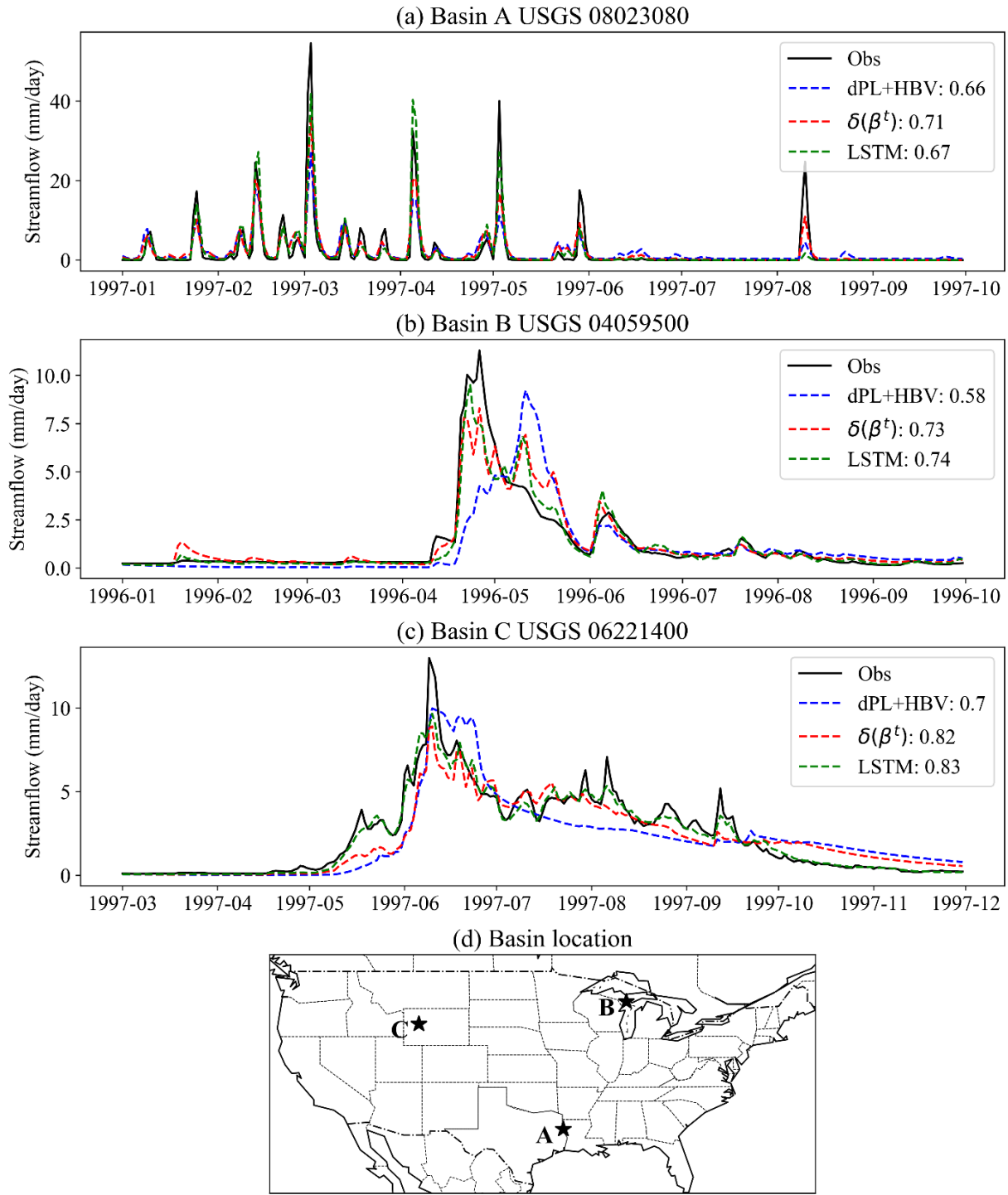

Figure 3. (a-c) Time series comparisons for several basins with NSE values close to the median -- comparing to dPL+original HBV vs. LSTM vs evolved HBV with DP, $\delta(\beta^t)$. The numbers in the legends show the NSE metric for the whole testing period (1995-2010). (d) Locations of basins in (a-c).

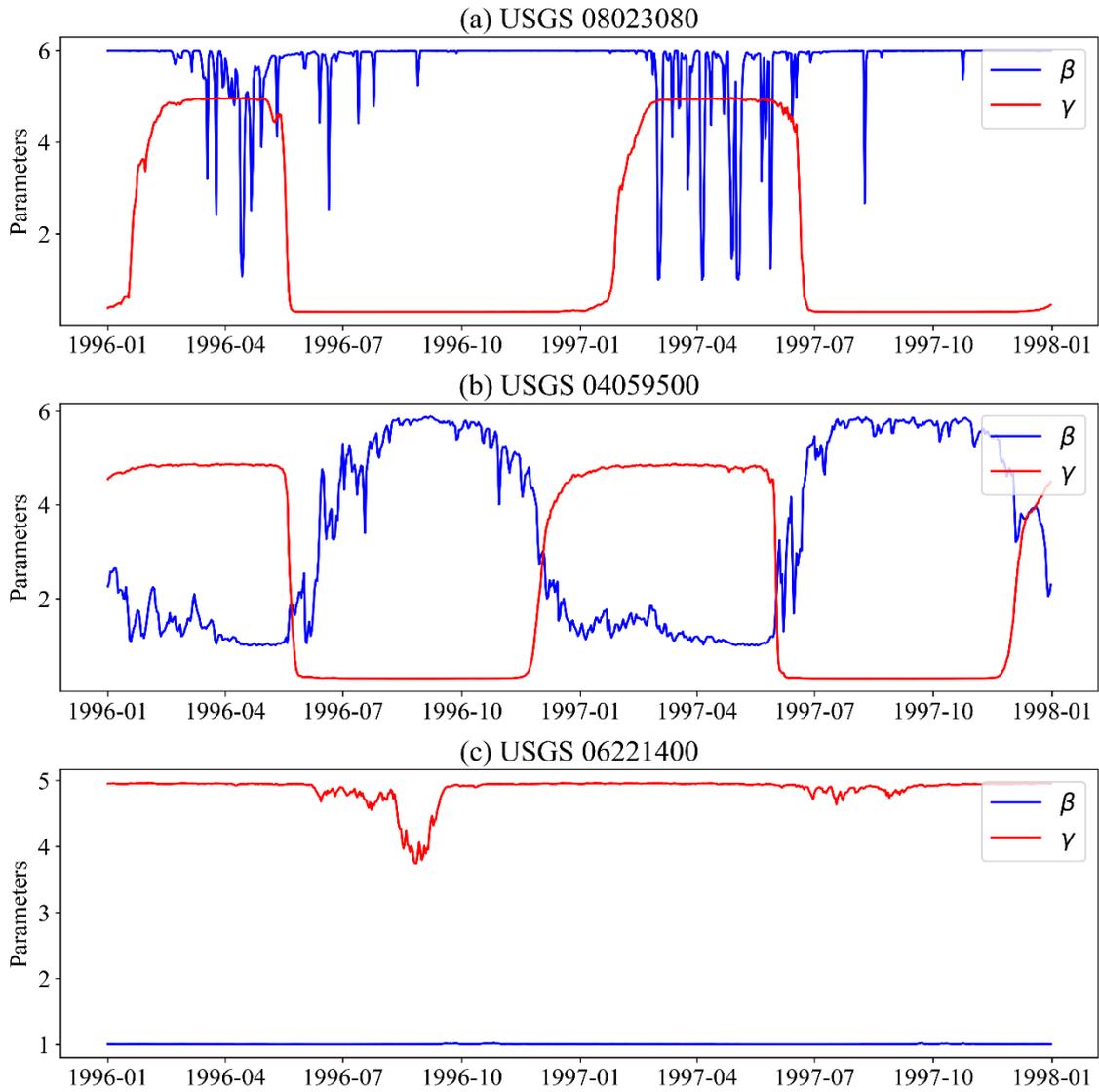

*Figure 4. Time series of dynamical parameters β and γ estimated from the model δ(β$^t$) and δ(γ$^t$), respectively, for the three basins shown in Figure 3.*

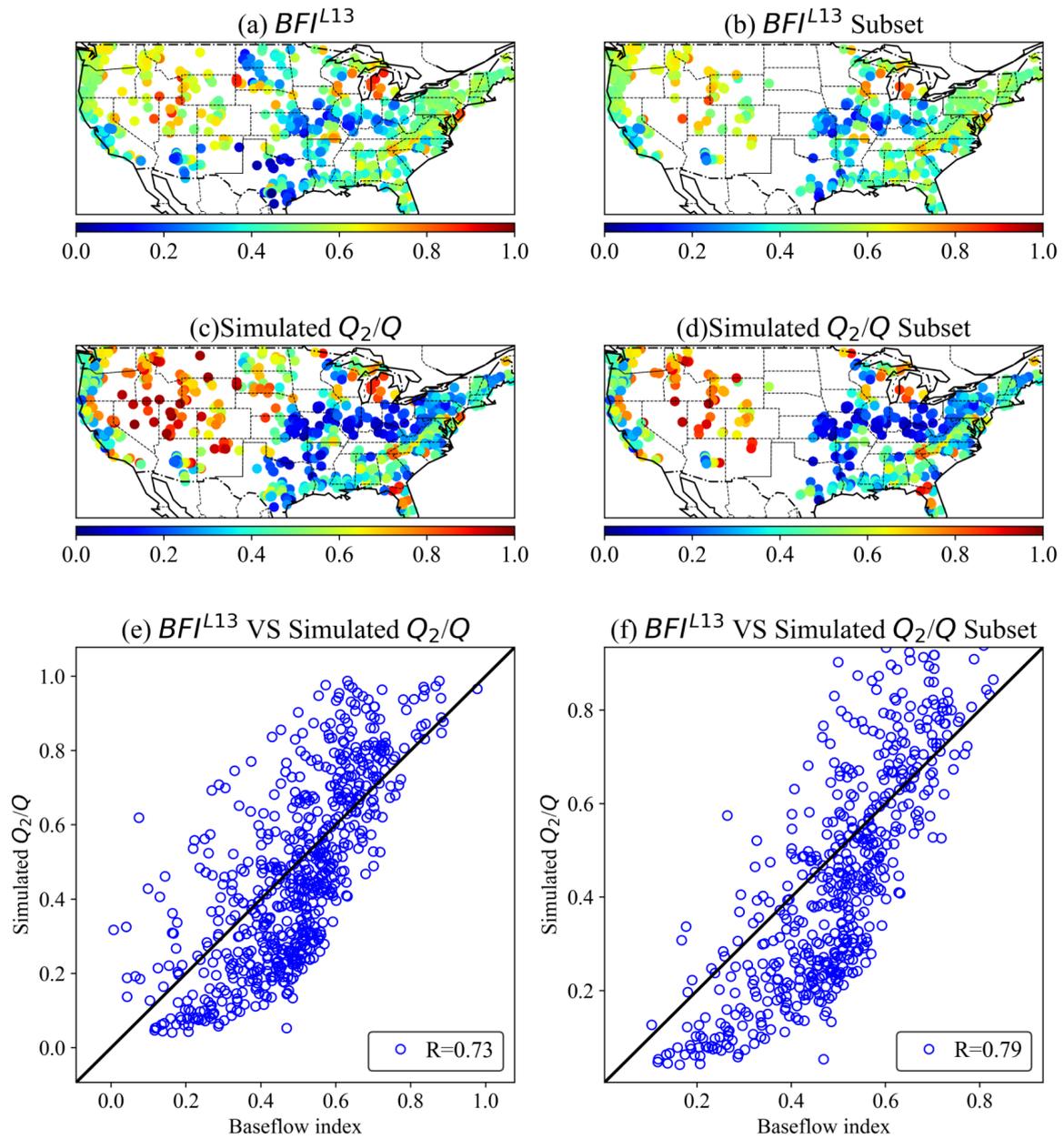

*Figure 5. (a) Map of baseflow index (BFI) according to USGS baseflow separation analysis; (c) Baseflow predicted by the trained model (long-term average $Q_2/Q$ from evolved HBV); (e) the correlation between USGS BFI and $Q_2/Q$. (b,d,f) Same as (a,c,e) but for the subset of basins with NSE>0.5.*

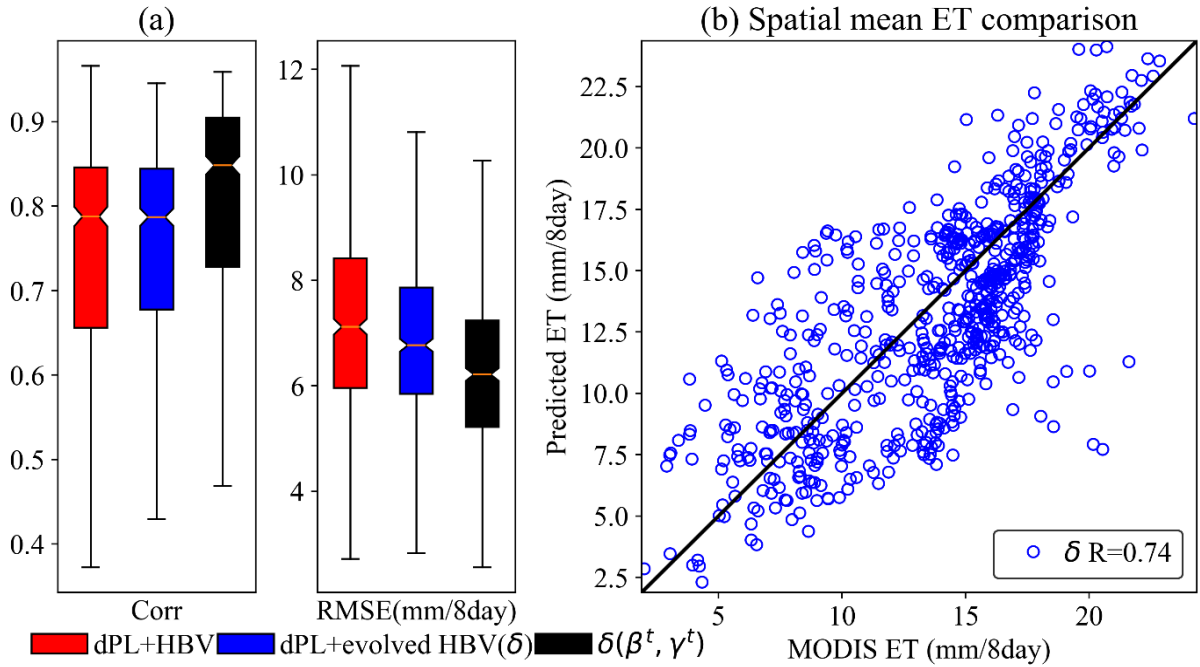

Figure 6. (a) Temporal correlation between simulated ET and the estimates from MOD16-MT 8-day product for the CAMELS basins (each box summarizes 671 basins), from three dPL HBV models; (b) Spatial correlation of the long-term mean ET (each circle represents a basin).

3.3. Further discussion

In this work we limited ourselves to frameworks that support prediction in ungauged basins, even though we leave the rigorous PUB benchmark to future work. Our parameterization and learning schemes are regionalized and based on widely-available inputs, and are thus applicable to large scales. Due to the lack of physical laws, LSTM may not learn the true causal relationship between static inputs and outputs. Therefore, we hypothesize the decline in performance from training basins to ungauged basins will be less significant for our learnable physical model than LSTM. Our future work will rigorously evaluate this hypothesis.

It is an expert's choice which model we use as a backbone and how much model structure we retain, but that choice impacts what kind of question can be asked. Here we chose a backbone that discerns soil moisture, groundwater, quickflow, and baseflow so we can support downstream applications like stream temperature and ecosystem modeling. On the flip side, one could use a much more complex model as a backbone, but generally, adding too many structural constraints tends to degrade model accuracy. In the end, we may have to make a conscious choice between interpretability and accuracy. We recommend backbone models with a process granularity that enables providing a narrative to stakeholders.

Having process granularity and physical fluxes and variables gives us another advantage --- we can now use multiple observations to constrain the model or inform unobserved variables. For example, since soil moisture observations are more widely available from satellite missions, we could use these observations, apart from streamflow, to constrain the model and update model states using the DL equivalent of data assimilation. Including additional observations to constrain different parts of the hydrologic model could reduce uncertainty and make the model more robust (Dembélé et al., 2020; Efstratiadis & Koutsoyiannis, 2010). This would not be possible for an alternative model whose backbone does not include soil moisture as an output.

Here the NNs are embedded into a time-discrete model which explicitly integrates over the time steps. Strictly speaking, the NNs in this framework learn the time-integrated operator which varies based on the time step size. This could be interpreted as using a forward Euler scheme to integrate over time. It is also possible to place the NN on the differential equation and use more accurate numerical solvers for time integration (Rackauckas et al., 2021), which would force the NN to learn the continuous operator. Both approaches are valid. The discrete version would require little change to the existing models (the code could be translated verbatim) which already have well-understood numerical schemes. The differential equation version would need stable integrator schemes in place, but it would also be a valuable avenue to explore.

While we showed the necessity of employing dynamical parameterization (DP) to truly stay on par with LSTM, we caution against an overuse of DP in the model. For one, many parameters, such as groundwater recession parameters, should not be dynamic. For two, because LSTM is so powerful, if we apply DP throughout the system, it has a high chance of achieving high performance, but we may start to lose physical significance and the system may revert to being an LSTM variant. The effects of DP should be quantified, its role justified by physical mechanisms, and its importance should be limited. In our case, DP only had a minor impact (median NSE 0.697 to 0.715). In other words, the variance to be explained by DP should not overwhelm the physical part. For future differentiable modeling methods, one important research direction would be system ways to limit or constrain the roles of neural networks and retain physical significance while allowing the framework to adapt to data.

We think the groundwater component still has significant room for improvement because the current groundwater formulations are too simple. Future efforts should pay special attention to improving the groundwater component, even though groundwater improvement does not necessarily get reflected in the NSE values.

## 4. Conclusion

The strong performance of the evolved HBV model shows that generalist model architecture like LSTM is not entirely necessary to achieve high performance. The success of this framework means we can now use differentiable programming to ask mechanistic questions. The above statements are not to say pure DL models are not useful -- they remain highly valuable because they can be quickly set up to gauge the information content in datasets, and they can provide a wealth of diagnostic signals.

We have imposed upon ourselves some stringent conditions (mass conservation, physical calculations, and interpretable physical outputs). The unchanged parts of the numerical model serve as physical constraints, allowing our model to output physical states and fluxes that are valuable for downstream applications. Utilizing a backbone with a sufficient level of process granularity is important to help with this mission.

The strengths of the differentiable models and LSTM models over traditional hydrologic models highlight the power of adaptive learning capacity. There seems to be a chasm between the performance of models with or without learnable components, but the gaps are minor between those alternatives that do have them. However, to enable learning complex functions, differentiable programming will be required because traditional optimization capabilities can barely handle more than 20 parameters. Numerically approximating the derivatives is prohibitively expensive for large neural networks. Thus, we perceive differentiable programming as an inevitable and promising path toward substantial growth.

Since the evolved HBV model has better results than dPL+HBV, it is now easier (compared to using LSTM) to answer the questions, e.g., *What should have been the equation if the original one was not good enough*? We only conducted some preliminary studies in this work, but hopefully it should be obvious that this flexibility allows us to open a whole new line of research. There are exciting opportunities to leverage this framework for asking novel questions or addressing some of hydrologists' nemesis questions in the near future.

## Appendix

*Table A1 The summary of input variables to the parameter learning neural network including 3 dynamic forcings and 35 static attributes.*

|  | Variables | Descriptions | Units |
| --- | --- | --- | --- |

| Forcing | P | Precipitation | mm/day |
|---|---|---|---|
| | T | Daily mean temperature | °C |
| | $E_p$ | Potential evapotranspiration | mm/day |
| Attributes | p_mean | Mean daily precipitation | mm/day |
| | pet_mean | Mean daily PET | mm/day |
| | p_seasonality | Seasonality and timing of precipitation | - |
| | frac_snow | Fraction of precipitation falling as snow | - |
| | aridity | PET/P | - |
| | high_prec_freq | Frequency of high precipitation days | days/year |
| | high_prec_dur | Average duration of high precipitation events | days |
| | low_prec_freq | Frequency of dry days | days/year |
| | low_prec_dur | Average duration of dry periods | days |
| | elev_mean | Catchment mean elevation | m |
| | slope_mean | Catchment mean slope | m/km |
| | area_gages2 | Catchment area (GAGESII estimate) | km$^2$ |
| | frac_forest | Forest fraction | - |
| | lai_max | Maximum monthly mean of the leaf area index | - |
| | lai_diff | Difference between the maximum and minimum monthly mean of the leaf area index | - |
| | gvf_max | Maximum monthly mean of the green vegetation | - |
| | gvf_diff | Difference between the maximum and minimum monthly mean of the green vegetation fraction | - |
| | dom_land_cover_frac | Fraction of the catchment area associated with the dominant land cover | - |
| | dom_land_cover | Dominant land cover type | - |
| | root_depth_50 | Root depth at $50^{th}$ percentiles | m |
| | soil_depth_pelletier | Depth to bedrock | m |
| | soil_depth_statgso | Soil depth | m |
| | soil_porosity | Volumetric soil porosity | - |
| | soil_conductivity | Saturated hydraulic conductivity | cm/hr |
| | max_water_content | Maximum water content | m |
| | sand_frac | Sand fraction | % |
| | silt_frac | Silt fraction | % |
| | clay_frac | Clay fraction | % |
| | geol_class_1st | Most common geologic class in the catchment | - |
| | geol_class_1st_frac | Fraction of the catchment area associated with its most common geologic class | - |
| | geol_class_2nd | Second most common geologic class in the catchment | - |

| | geol_class_2nd_frac | Fraction of the catchment area associated with its 2nd most common geologic class | - |
| | carbonate_rocks_frac | Fraction of the catchment area as carbonate sedimentary rocks | - |
| | geol_porosity | Subsurface porosity | - |
| | geol_permeability | Subsurface permeability | $m^2$ |

## Acknowledgments

DF was supported by US National Science Foundation Award EAR #1832294. CS, JL and KL were supported by US National Science Foundation Award OAC #1940190. Computing was partially supported by US National Science Foundation Award PHY #2018280. Some results have been presented in the AI4ESP conference in Nov 2021 (https://youtu.be/kC2JyvaUd58?t=410) and the American Geophysical Union (AGU) Annual Fall Meeting Dec 2021, among other seminars. An earlier version of the manuscript has also been uploaded to arxiv. According to AGU policy, these presentations are not considered dual publications.

The deep learning code relevant to this work can be downloaded at http://doi.org/10.5281/zenodo.5015120. The new HBV training code will be made available upon a revision request, and a new Zenodo release will be published upon paper acceptance. CAMELS data can be downloaded at https://ral.ucar.edu/solutions/products/camels. The extended NLDAS forcing data for CAMELS can be downloaded at https://www.hydroshare.org/resource/0a68bfd7ddf642a8be9041d60f40868c/. MODIS ET data can be downloaded at https://modis.gsfc.nasa.gov/data/dataprod/mod16.php.

spatial patterns with multiple satellite data sets. *Water Resources Research*, *56*(1), e2019WR026085. https://doi.org/10.1029/2019WR026085

Dick, J. J., Soulsby, C., Birkel, C., Malcolm, I., & Tetzlaff, D. (2016). Continuous Dissolved Oxygen Measurements and Modelling Metabolism in Peatland Streams. *PLOS ONE*, *11*(8), e0161363. https://doi.org/10.1371/journal.pone.0161363

Efstratiadis, A., & Koutsoyiannis, D. (2010). One decade of multi-objective calibration approaches in hydrological modelling: A review. *Hydrological Sciences Journal*, *55*(1), 58–78. https://doi.org/10.1080/02626660903526292

Fang, K., Kifer, D., Lawson, K., Feng, D., & Shen, C. (2022). The data synergy effects of time-series deep learning models in hydrology. *Water Resources Research*. https://doi.org/10.1029/2021WR029583

Fang, K., Pan, M., & Shen, C. (2019). The value of SMAP for long-term soil moisture estimation with the help of deep learning. *IEEE Transactions on Geoscience and Remote Sensing*, *57*(4), 2221–2233. https://doi.org/10/gghp3v

Fang, K., & Shen, C. (2017). Full-flow-regime storage-streamflow correlation patterns provide insights into hydrologic functioning over the continental US. *Water Resources Research*, *53*(9), 8064–8083. https://doi.org/10.1002/2016WR020283

Fang, K., & Shen, C. (2020). Near-real-time forecast of satellite-based soil moisture using long short-term memory with an adaptive data integration kernel. *Journal of Hydrometeorology*, *21*(3), 399–413. https://doi.org/10/ggj669

Fang, K., Shen, C., Kifer, D., & Yang, X. (2017). Prolongation of SMAP to spatiotemporally seamless coverage of continental U.S. using a deep learning neural network. *Geophysical Research Letters*, *44*(21), 11,030-11,039. https://doi.org/10/gcr7mq

Feng, D., Fang, K., & Shen, C. (2020). Enhancing streamflow forecast and extracting insights using long-short term memory networks with data integration at continental scales. *Water Resources Research*, *56*(9), e2019WR026793. https://doi.org/10.1029/2019WR026793
32